\documentclass[preprint,twocolumn,5p,12pt]{article}

\usepackage{amssymb}
\usepackage{subcaption}
\usepackage{graphicx}
\usepackage{caption}
\usepackage{xspace}
\usepackage{multirow}
\usepackage{rotating}
 \usepackage{url}
\usepackage{gensymb}
\usepackage [ table ]{ xcolor }
\usepackage[top=2cm, bottom=2cm, left=1.5cm, right=1.5cm]{geometry}
\usepackage{setspace}
\usepackage{morefloats}

\usepackage{commath}

\usepackage[affil-it]{authblk}
\usepackage[colorlinks=true,citecolor=blue, urlcolor=blue, linkcolor=blue]{hyperref}

\usepackage{setspace}

\title{Spike time displacement based error backpropagation in convolutional spiking neural networks}
\author{Maryam Mirsadeghi$ ^{1}$}
\author{Majid Shalchian$ ^{1,}$\footnote{Corresponding Author \\Email addresses: \href{mailto://m.mirsadeghi@aut.ac.ir}{m.mirsadeghi@aut.ac.ir} (MM), \\ \href{mailto:// shalchian@aut.ac.ir}{ shalchian@aut.ac.ir} (MM), \\
\href{mailto:// s\_kheradpisheh@sbu.ac.ir}{ s\_kheradpisheh@sbu.ac.ir} (SRK), \\
\href{mailto:/timothee.masquelier@cnrs.fr}{timothee.masquelier@cnrs.fr} (TM)} }
\author{Saeed Reza Kheradpisheh$ ^{2}$}
\author{\\Timoth\'ee Masquelier$ ^{3}$}
\affil{\footnotesize $ ^{1} $ Department of Electrical Engineering, Amirkabir University of Technology, Tehran, Iran}
\affil{\footnotesize $ ^{2} $ Department of Computer and data Sciences, Shahid Beheshti University, Tehran, Iran}
\affil{\footnotesize $ ^{3} $  CerCo UMR 5549, CNRS Universit\'e Toulouse 3, France}

\date{}

\usepackage[absolute]{textpos}
\begin{document}
\maketitle

\begin{abstract}
We recently proposed the STiDi-BP algorithm, which avoids backward recursive gradient computation, for training multi-layer spiking neural networks (SNNs) with single-spike-based temporal coding. The algorithm employs a linear approximation to compute the derivative of the spike latency with respect to the membrane potential and it uses spiking neurons with piecewise linear postsynaptic potential to reduce the computational cost and the complexity of neural processing.
In this paper, we extend the STiDi-BP algorithm to employ it in deeper and convolutional architectures. The evaluation results on the image classification task based on two popular benchmarks, MNIST and Fashion-MNIST datasets with the accuracies of respectively $99.2\%$ and $92.8\%$, confirm that this algorithm has been applicable in deep SNNs.
Another issue we consider is the reduction of memory storage and computational cost.
To do so, we consider a convolutional SNN (CSNN) with two sets of weights: real-valued weights that are updated in the backward pass and their signs, binary weights, that are employed in the feedforward process.
We evaluate the binary CSNN on two datasets of MNIST and Fashion-MNIST, and obtain acceptable performance with a negligible accuracy drop with respect to real-valued weights (about $0.6\%$ and $0.8\%$ drops, respectively).
\end{abstract}


\section{Introduction}
Spiking neural networks (SNNs) are recently attracting more and more attention due to their temporal nature and their event-driven processing paradigm which make them suitable for energy-efficient neuromorphic implementation.
However, due to the use of non-differentiable activation function and the temporal dynamics of SNNs, it is still a big challenge to train SNNs directly. Therefore, they have not yet reached the state-of-the-art accuracy compared to the artificial neural networks (ANNs), especially, in deep architectures with single-spike-based temporal coding. 

In temporal coding scheme, information is carried by the timing or the order of individual spikes~\cite{K1,K2,K3}. In the extreme case of single-spike-based temporal coding, neurons are allowed to fire at most once, which can radically reduce the computational and energy demand of SNNs.
So far, different solutions have been proposed to adapt the backpropagation (BP) algorithm to directly train SNNs with single-spike-based temporal coding.

Since the neuronal activity in single-spike coding is defined by the neurons firing time, two approaches are used to adapt BP to single-spike-based SNNs. The first approach is to compute or approximate the derivative of the firing time of each neuron with respect to its membrane potential~\cite{R7, R10, R18}. The second approach is to directly compute the firing time of each postsynaptic neuron based on the firing times of its presynaptic neurons~\cite{R8, R9, S1}.

Bohte, et al.~\cite{R7} introduced a temporal version of BP called SpikeProp which minimizes the temporal error of the network to train single-spike multilayer SNNs. They used exponentially SRM neuron models and employed a piecewise linear approximation to compute the derivative of the thresholding activation function at the firing time.
Kheradpisheh, et al.~\cite{R10} proposed temporal version of BP for a multi-layer SNN  with IF neurons and instantaneous synapses. To do so, they approximated the derivative of the neurons firing latency with respect to the  membrane potential by $-1$. By using  Rectified Linear Postsynaptic Potential (ReL-PSP) spiking neuron model, Zhang et al.~\cite{R18} could avoid such approximation and precisely compute this derivative. Mostafa~\cite{R8} defined the firing time of each neuron directly based on its presynaptic spike times. He used IF neurons with exponentially
decaying synaptic current.  Comsa, et al.~\cite{R9} employed a similar approach for SRM neuron models with alpha synaptic function. 
Zhou et al.~\cite{S1} developed \cite{R8} for implementing deep convolutional spiking neural networks (CSNNs) and achieved the state-of-the-art performance.

All aforementioned models but Zhang et al~\cite{R18} and Zhou et al.~\cite{S1}, have been used fully-connected networks with one or two hidden layers and they have not ever been applied to a deeper structure. In Zhang et al.~\cite{R18}, authors developed a deep convolutional spiking neural network consisted of two convolutional and two hidden layers~\cite{R18}. They used ReL-PSP based spiking neuron model and trained the network by employing temporal BP with recursive backward gradient. 
Zhou et al.~\cite{S1} extended \cite{R8} to implement deep architectures of SNN based on the VGG16 model~\cite{S2,S3} for CIFAR10 and the GoogleNet model~\cite{S4} for ImageNet.
To the best of our knowledge, These are the only implementation of a CSNN  with single-spike-based temporal coding. Other CSNNs are either a converted version of traditional CNNs~\cite{R4, R5, R6, R46} or they use rate coding or multi-spike per neuron schemes to directly apply BP on the network~\cite{R11,R12,R13}.

Recently, we proposed an error backpropagation algorithm based on the spike time displacements, called STiDi-BP, for multi-layer fully-connected SNNs with single-spike-based temporal coding~\cite{R3}. Similar to~\cite{R18, R7, R10}, we used a linear approximation to compute the derivative of neurons' firing time with respect to their membrane potential. However, in STiDi-BP, instead of recursively backpropagating the errors, we computed the desired firing time of neurons in each layer, and hence, we could locally compute the error just by comparing the actual and desired firing times. 

In this paper, we extend the STiDi-BP learning approach to be applicable in deeper and convolutional architectures. The evaluation results on two image classification tasks of MNIST and Fashion-MNIST datasets, with respectively $99.2\%$ and $92.8\%$ recognition accuracy, confirm the capabilities of the proposed algorithm in deep CSNNs.

Implementing SNNs with real-valued weights on neuromorphic devices requires a large amount of memory space and imposes a high load of floating-point computation. Binarizing the synaptic weights can help to reduce the memory footprint and the computational cost.
A few recent studies had tried to convert supervised binary artificial neural networks (BANNs) into equivalent binary SNNs (BSNNs)~\cite{R14,R15,R16,R19}. For the first time, Kheradpisheh et al. have introduced a direct supervised learning algorithm to train a two-layer fully connected SNN with binary synaptic weights\cite{R17}. But they didn't apply it to deeper SNNs or CSNNs. There is no other study to the best of our knowledge aimed at directly training deep supervised SNNs and CSNNs with binary weights. 
Here, we employ STiDi-BP to directly train a deep CSNN with binary synaptic weights which are the sign of real-valued weights. In the backward pass, we update the real-valued weights and the feedforward processing is performed by the binary weights.
We have evaluated the proposed network on MNIST and Fashion-MNIST datasets with categorization accuracies of $98.6\%$ and $92.0\%$, respectively, that has a negligible drop compared to real-valued-based CSNN.

\section{Forward pass}

Here, the proposed convolutional spiking neural network is comprised of a temporal coding input layer, a stack of interlaying convolutional and pooling layers for feature exraction, and a cascade of fully connected layers for the final classification. 
A temporal coding is used to convert the input image into a sparse spike train (i.e. one spike per pixel). After feeding the input codded image to the network, the convolutional operations are applied. 
In a convolutional layer, several filters are used to extract visual features from the previous layer which are presented in different feature maps. 
After each convolutional layer, a pooling layer is used to remove the redundancy and reduce the size of the feature maps.
A pooling layer does a nonlinear max pooling operation over a set of neighboring neurons to select a neuron with the highest activity (i.e., earliest spike).
After the last pooling layer, the fully connected layers are implemented to process the extracted features and do the final classification.

\subsection{Neuron Model}
\begin{figure}
\centering
\includegraphics[width=0.5\textwidth]{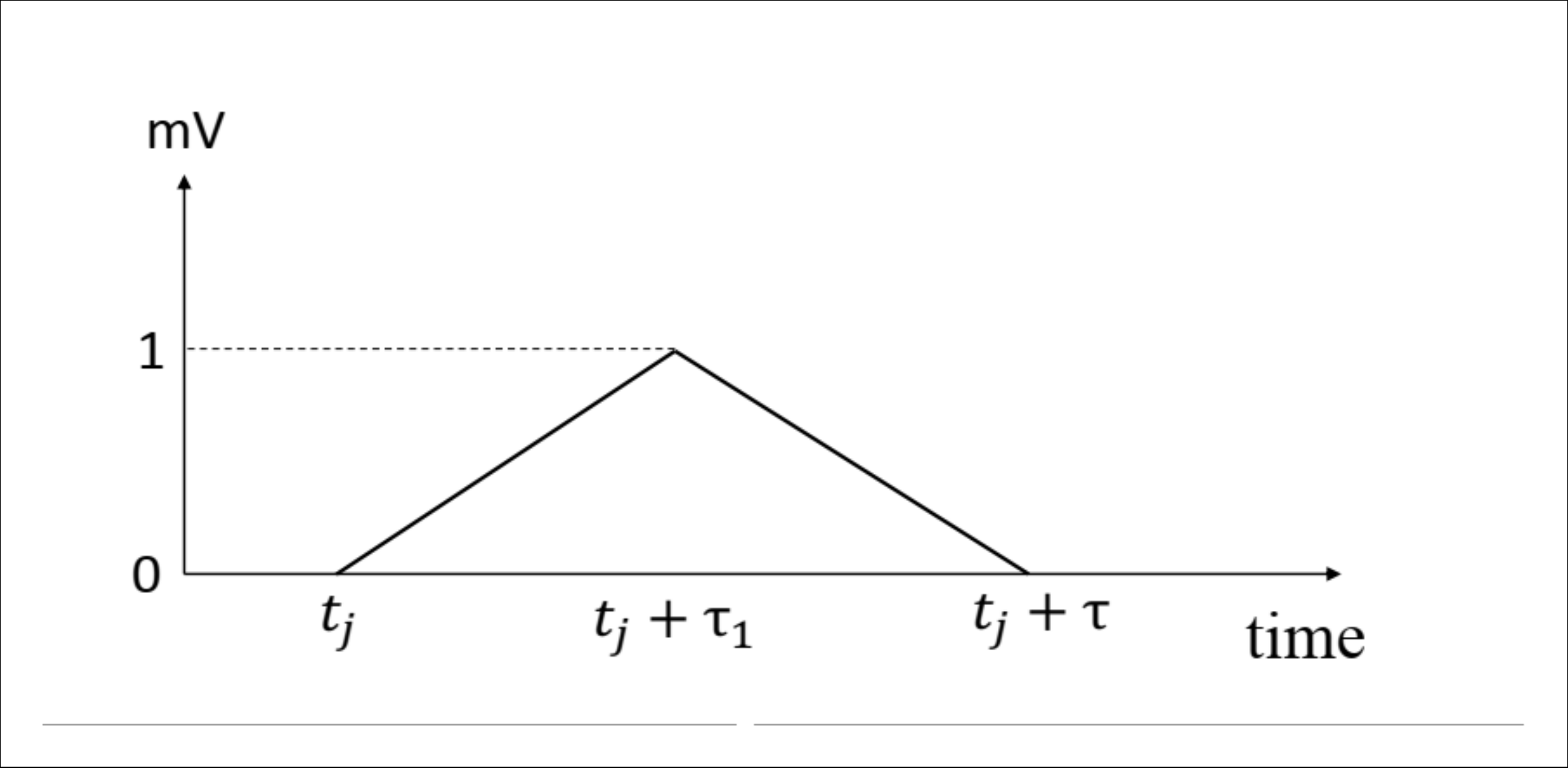} 
\caption{The piecewise linear postsynaptic potential caused by the presynaptic spike time $t_j$.}
\label{FIG1}
\end{figure}

We use a simple piecewise linear postsynaptinc potential
(PL-PSP) based spiking neuron model which has a very low computational cost compared to exponential PSP models~\cite{R3}. The membrane potential $v_i(t)$ of neuron $i$
at time $t$ is the weighted summation of the PL-PSPs of its afferent neurons:
\begin{equation}
v_i(t) = \sum\limits_{j\in{J}} w_{ij} \epsilon(t-t_j),    
\label{eq1}
\end{equation}
where, $w_{ij}$ is the synaptic weight connecting the presynaptic neuron $j$ to the neuron $i$ and $t_j$ is the spike time of neuron $j$. $\epsilon(t-t_j)$ is the kernel of the PL-PSP function that is illustrated in Figure~\ref{FIG1} and is described by the following equation:

\begin{equation}
\epsilon(t-t_j) =
\begin{cases}
\frac{t-t_j}{\tau_1}&\text{if }\ t_j \leq t <t_j + \tau_1,\\
\frac{t_j+\tau-t}{\tau_2}&\text{if }\ t_j + \tau_1 \leq t <t_j + \tau,\\
\end{cases}
\label{eq2}
\end{equation}
here, $\tau_1$ and $\tau_2$ are time constants of the PL-PSP function and $\tau=\tau_1+\tau_2$.

\subsection{Temporal coding}
Contrary to the costly rate-coding scheme in which the input image is encoded in the spike rates of the input neurons (i.e., the higher the pixel value, the higher the firing rate), we use the more efficient temporal coding scheme\cite{R3}, where the information is carried by the timing of individual spikes in a sparse manner. Each input neuron fires at most once such that neurons corresponding to pixels with higher intensities emit earlier spikes.
After feeding input spikes to the network, each neuron in the subsequent layer updates its membrane potential by integrating the voltage sum of all the presynaptic spikes and fires a spike right after crossing the threshold. Each neuron fires only once where its firing time determines the saliency of the extracted feature (such as the input layer), hence, the whole network obeys the sparse temporal coding.

\subsection{Convolutional layers}
There are several feature maps in a convolutional layer, each of which corresponds to a convolutional filter.
Neurons in a specific map share the same set of synaptic weights, and therefore, detect the same feature at different locations. 
Within a map $D$, each convolutional neuron at location ($i$,$j$) receives spikes from neurons inside a certain window within all the feature maps of the previous layer. 
Therefore, each visual feature in a convolutional layer is obtained by combining several simpler features extracted in the previous layer.
The neuron computes its membrane potential $v_{ij}^D$ by applying the corresponding filter $d$ on the received spike times according to the extended Eq.~\ref{eq1} and Eq.~\ref{eq2}.
Whenever $v_{ij}^D$ crosses the threshold $v_{th}$, the convolutional neuron emits a spike, and regarding the single-spike-based coding, it will remain silent until the end of the simulation.

\subsection{Pooling layers}

In the rate-coding scheme, it is not possible to detect the neuron with maximum firing rate until the last simulation time point. 
While, in the proposed model, due to the use of temporal coding, we can simply perform the max pooling operation by propagating the first spike appearing in the input window of each pooling neuron.
To do so, in pooling layers, we use IF neurons with the threshold and the input synaptic weights of one. 
Each pooling neuron performs the maximum operation over a window in the corresponding feature map of the previous layer.
The first input spike from the neighboring afferent neurons activates the pooling neuron and makes it to fire a spike immediately. Each pooling neuron is permitted to fire at most once during the simulation time. 
Note that there is no learning for the pooling neurons.

\subsection{Fully connected layers}

The last stage of the proposed model which performs the final classification based on the extracted visual features of the convolutional part, is composed of cascading fully connected layers with PL-PSP based neurons.

Each neuron $i$ in layer $l$ integrates the weighted spikes according to Eq.~\ref{eq1} and Eq.~\ref{eq2} and emits a spike when its post-synaptic voltage $v_i^l$ crosses the threshold. The neuron is allowed to fire at most once and the spike latency carries the information.
The same process has recurred in the following hidden layers and output layers.

In the output layer, the number of neurons is equal to the number of classes and each neuron is assigned to a different category. The output neuron that fires earlier than others is called the winner and it determines the class of the input image.

\section{Backward pass}

Here we apply the proposed learning algorithm to a deep convolutional spiking neural network to show that it is suitable and practical for deep structure of SNNs.

The error loss function of each layer is described independently by computing the time differences between the actual and the desired firing times. The desired firing times in the middle layers are calculated by displacing the presynaptic spike times such that the error is minimized.
Then, the GD is performed locally to update the synaptic weights without any backward recursive GD computation and to overcome the non-differentiality of SNNs, a linear approximation method described in \cite{R3} is employed.

\subsection{Fully connected layers}
The learning rule in the fully connected layers is the same as~\cite{R3}.
Here, we briefly introduce the proposed learning algorithm for the reader's convenience. More complete description of STiDi-BP is given in\cite{R3}.

The loss function of each layer $l$ is calculated independently by the following equation:
\begin{equation}
E^{l} = \sum\limits_{j}E_j^l = \sum\limits_{j}\frac{1}{2}(e_{j}^{l})^2,
\label{eq3}
\end{equation}
where, $e_{j}^{l}$ is the temporal error function for the postsynaptic neuron $j$ obtained by substracting the desired and the actual firing times ($T_{j}^l$ and $t_{j}^l$, respectively) of the neuron $j$ in the $l^{th}$ layer:
\begin{equation}
e_{j}^{l} = \frac{T_{j}^l-t_{j}^l}{T_{max}}.
\label{eq4}
\end{equation}

In order to minimize the squared error loss function $E^l$, the synaptic weights of layer $l$ should be modified by using GD algorithm. 
To update each synaptic weight $w_{ji}^l$, we compute the gradient of loss function with respect to $w_{ji}^l$. Hence
\begin{equation}
w_{ji}^l(new) = w_{ji}^l(old) + \Delta w_{ji}^l = w_{ji}^l - \eta \frac{\partial E^l}{\partial w_{ji}^l},
\label{eq5}
\end{equation}
where, $\eta$ is the learning rate parameter and $w_{ji}^l$ is the weight connection between neuron $i$ in layer $l-1$ to the neuron $j$.
$\Delta w_{ji}^l$ in Eq.~\ref{eq5} can be expanded to
\begin{equation}
\begin{cases}
\Delta w_{ji}^l = -\eta \frac{\partial E_j^{l}}{\partial t_{j}^l}\frac{\partial t_{j}^l}{\partial v_{j}^l(t)}\frac{\partial v_{j}^l(t)}{\partial w_{ji}^l},&\text{if}\ t_{i}^{l-1} \leq t_{j}^l \\
0,&\mathrm{otherwise}
\end{cases}
\label{eq6}
\end{equation}
where, $v_{j}^l(t)$ is the membrane potential of neuron $j$ and $t_{i}^{l-1}$ is the spike time of neuron $i$. Neuron $h$ has contribution to the membrane potential computation only if it fires before $t_j^l$.
By using Eq.~\ref{eq3} and Eq.~\ref{eq4} we can express the first term as:
\begin{equation}
\frac{\partial E_j^{l}}{\partial t_{j}^l} = -\frac{T_j^l - t_{j}^l}{T_{max}^2} = -\frac{e_j^{l}}{T_{max}},
\label{eq7}
\end{equation}
And, the third term is computed by considering Eq.~\ref{eq1} and Eq.~\ref{eq2}:
\begin{equation}
\frac{\partial v_{j}^l(t)}{\partial w_{ji}^l} = \epsilon(t_{j}^l - t_{i}^{l-1}).
\label{eq8}
\end{equation}

The second term, the derivative of the postsynaptic spike time with respect to its membrane potential, is calculated according to the following equation.
The details of computation are given in \cite{R3}.
\begin{equation}
\frac{\partial t_{j}^l}{\partial v_{j}^l(t)} = -\frac{1}{\frac{\partial v_{j}^l(t)}{\partial t_{j}^l}} \simeq -\frac{t_{j}^l}{v_{th}}.
\label{eq9}
\end{equation}

By substituting Eq.~\ref{eq7}, Eq.~\ref{eq8} and Eq.~\ref{eq9} in Eq.~\ref{eq6}, the final equation for modifing the synaptic weights and minimizing the squared error loss function $E^l$ is described by:
\begin{equation}
\begin{cases}
\Delta w_{ji}^l = -\eta\frac{e_j^{l}}{T_{max}}.\frac{t_{j}^l}{v_{th}}.\epsilon(t_{j}^l - t_{i}^{l-1}),&\text{if}\ t_{i}^{l-1} \leq t_{j}^l \\
0&\mathrm{otherwise}
\end{cases}
\label{eq21}
\end{equation}

After computing the error loss function for each layer, we use Eq.~\ref{eq5} to update the synaptic weights of that layer and minimize the local error. Hence, we don't have backward recursive gradient computation. 
How to calculate target firing times is an important issue that will be discussed in the next section.

\subsubsection{Calculation of target firing times}
The target firing time is computed using different formula for the neurons of middle layer and output layer~\cite{R3}. 

In the output layer, we use a relative encoding method in which the correct output neuron should be encouraged to fire earlier than others. 
To do that, should take into account the input image category lablel. 
by assuming that the input image belongs to the $i^{th}$ class, the $i^{th}$ output neuron should fire at time $\tau_{min}-\lambda$, and others set to fire at later time $\tau_{max}+\lambda$.
Here $\tau_{max}$ and $\tau_{min}$ are the maximum and the minimum output firing times and $\lambda$ is a constant parameter used to provide resolution distance for the winner neuron.

There is a different situation for the middle layers. 
To compute the desired firing time of each neuron $j$ in the $l^{th}$ middle layer, we define the time displacement amount of the neuron spike time $t_j^l$ to reduce the postsynaptic error $E^{l+1}$. 
To do that, we compute the derivative of the postsynaptic error with respect to $t_j^l$:
\begin{equation}
\Delta t_j^{l} = -\beta \sum\limits_{i}\frac{\partial E_i^{l+1}}{\partial t_j^{l}},    \\t_{j}^{l} \leq t_{i}^{l+1}.
\label{eq17}
\end{equation}
Here, $\beta$ is the learning rate and $j$ iterates over neurons in layer $l+1$.
By expanding the Eq.~\ref{eq17} we have
\begin{equation}
\Delta t_j^{l} = -\beta \sum\limits_{i} \frac{\partial E_i^{l+1}}{\partial t_i^{l+1}}\frac{\partial t_i^{l+1}}{\partial v_i^{l+1}(t)}\frac{\partial v_i^{l+1}(t)}{\partial t_j^{l}}.
\label{eq18}
\end{equation}
The first and the second terms of Eq.~\ref{eq18} are calculated  the same as Eq.~\ref{eq7} and Eq.~\ref{eq8}, respectively and, the third term is expressed by considering Eq.~\ref{eq1} and Eq.~\ref{eq2}
\begin{equation}
dv=\frac{\partial v_i^{l+1}(t)}{\partial t_j^l} =
\begin{cases}
-\frac{w_{ij}^{l+1}}{\tau_1},&\text{if }\ t_j^l \leq t_i^{l+1} <t_j^l + \tau_1\\
\frac{w_{ij}^{l+1}}{\tau_2},&\text{if }\ t_j^l + \tau_1 \leq t_j^{l+1} <t_i^l + \tau\\
\end{cases}
\label{eq19}
\end{equation}

Finally, the time displacement amount of presynaptic neuron $j$ is described by
substituting  Eq.~\ref{eq7}, Eq.~\ref{eq8} and Eq.~\ref{eq19} in the RHS of Eq.~\ref{eq18}
\begin{equation}
\Delta t_{j}^l = -\beta\sum\limits_{i}\frac{e_i^{l+1}}{T_{max}}(\frac{t_i^{l+1}}{v_{th}}) dv.
\label{eq20}
\end{equation}
The postsynaptic error $E^{l+1}$ is reduced if the presynaptic neuron $j$ fires a spike at time $T_j^l=t_j^{l} + \Delta t_{j}^l$ instead of time $t_j^l$. Hence, $T_j^l$ should be considered as the target firing time of the neuron $j$.

\subsection{Convolutional layers}

For each specific map $D$ of a convolutional layer $C$, the error loss function is calculated independently, by integrating the mean squares of the difference between the actual and the desired firing times:
\begin{equation}
E_{D}^{C} = \sum\limits_{d}E_d^C = \sum\limits_{d}\frac{1}{2}(T_d^C-t_d^C)^2,
\label{eq30}
\end{equation}
where, d iterates over all the neurons of map $D$. 
Then, the GD algorithm is locally employed to modify the synaptic weights of the corresponding filter $D$ as follows:
\begin{equation}
\Delta W_D^C = -\eta_c \frac{\partial E_D^C}{\partial W_D^C} = -\eta_c \sum\limits_{d}\frac{\partial E_d^C}{\partial W_D^C}.
\label{eq31}
\end{equation}

Each synaptic weight of filter $D$ ($w^c_{D_{nij}}$) corresponds to the presynaptic neuron located at ($n,i,j$) with the spike firing latency of $t_{nij}$.
Therefore, $w^c_{D_{nij}}$ is updated as 
\begin{equation}
\Delta w^C_{D_{nij}} = -\eta_c \sum\limits_{d}\frac{\partial E_d^C}{\partial w^C_{D_{nij}}}.
\label{eq32}
\end{equation}

By expanding Eq.~\ref{eq32}, we have
\begin{equation}
\Delta w^C_{D_{nij}} = -\eta_c \sum\limits_{d}\frac{\partial E_d^C}{\partial t^C_d} \frac{\partial t^C_d}{\partial v^C_d} \frac{\partial v^C_d}{\partial w^C_{D_{nij}}}.
\label{eq33}
\end{equation}
The first, second and third terms of Eq.~\ref{eq33} are calculated by extending Eq.~\ref{eq7}, Eq.~\ref{eq8} and Eq.~\ref{eq9}, respectivly:
\begin{equation}
\frac{\partial E_d^C}{\partial t_d^C} = -\frac{T_d^C - t_d^C}{T_{max}^2},
\label{eq34}
\end{equation}

\begin{equation}
\frac{\partial t_d^C}{\partial v_d^C(t)} = -\frac{1}{\frac{\partial v_d^C(t)}{\partial t_d^C}} \simeq -\frac{t_d^C}{v_{th}},
\label{eq35}
\end{equation}

\begin{equation}
\frac{\partial v_d^C(t)}{\partial w_{D_{nij}}^C} = \epsilon(t_d^C- t_{nij}),
\label{eq36}
\end{equation}
where, $t_d^C$ is the spike firing latency of neuron $d$ in map $D$ of convolutional layer $C$.

\section{Binarization}
Here we apply some modification to STiDi-BP learning rule to directly train the CSNN with binary synaptic weights $\{-1,1\}$.

The only change in the forward path is the use of binary weights $W_b$ instead of real-valued weights, where $W_b=sign(W)$. Hence, the membrane potential of the postsynaptic neuron $i$ of layer $l$  in Eq.~\ref{eq1} is rewriten as
\begin{equation}
v_i^l(t) = \alpha^l \sum\limits_{j\in{J}} w^{l}_{b_{ij}} \epsilon(t-t_j).
\label{eq22}
\end{equation}
$j$ iterates over all presynaptic neurons and $\alpha^l$ is a shared scaling factor among all neurons of layer $l$. Each layer has its own scaling factor which should be updated in addition to the synaptic weights in the learning phase. The scaling factor is used to make sure that neurons cross the threshold.

In the backward pass, we have two sets of weights, real-valued weights and binary weights.
For each layer $l$, we update the real-valued weights by using Eq.~\ref{eq5} and Eq.~\ref{eq21} explained in section 3.2 and, update the scaling factor $\alpha^l$ as
\begin{equation}
\alpha^l = \alpha^l + \Delta \alpha^l = \alpha^l - \mu \frac{\partial E^l}{\partial \alpha^l},
\label{eq23}
\end{equation}
here $\mu$ is the learning rate parameter and $\frac{\partial E^l}{\partial \alpha^l}$ is calculated by using the following equation:
\begin{equation}
\frac{\partial E^l}{\partial \alpha^l} = \sum\limits_{i}\frac{\partial E_i^{l}}{\partial t_{i}^l}\frac{\partial t_{i}^l}{\partial v_{i}^l(t)}\frac{\partial v_{i}^l(t)}{\partial \alpha^l}.
\label{eq24}
\end{equation}

The first and second terms of Eq.~\ref{eq24} are calculated according to Eq.~\ref{eq7}  and Eq.~\ref{eq8}. For computing the third term we use Eq.~\ref{eq22}:
\begin{equation}
\frac{\partial v_{i}^{l}(t)}{\partial \alpha^l} = \sum\limits_{j\in{J}} w_{b_{ij}} \epsilon(t-t_j).
\label{eq25}
\end{equation}

\section{Experiments and results}

In this section we evaluate the proposed STiDi-BP training algorithm on deep structure of spiking neural network for two image classification tasks: MNIST dataset and Fashion-MNIST dataset. We develop a single-spike-based temporal convolutional SNN with piecewise linear SRM neurons and consumes two different modes: real-valued weights and binary weights. In the following, each network (CSNN and binary CSNN) is examined separately.

\subsection{ Real-valued weights}
\subsubsection{MNIST dataset}
The MNIST dataset~\cite{mnist} is the most popular benchmark for spiking neural networks. It comprises of $60000$ $28*28$ grayscale training images and $10000$ $28*28$ grayscale testing images.
To evaluate the proposed learning algorithm on the MNIST dataset, we develop an Real-valued weights-base CSNN (R-CSNN) with the structure of $28*28-40C5-P2-1000-10$, which consists of one convolutional layer, one pooling layer and one hidden layer followed by an output layer.
The convolutional layer is comprised of $40$ neural maps with $5*5$ convolution- window. The pooling-window of the pooling layer is of size $2*2$ with the stride of $2$. The hidden and the output layers are respectively consist of $1000$ and $10$ neurons.
Here the maximum simulation time is $T_{max}=100$ and other parameters of each layer are listed separately in Table~\ref{tab1}
 
\begin{table} 
\begin{center}
\caption{Model parameters for MNIST dataset in R-CSNN.}\label{tab1}
\resizebox{0.5\textwidth}{!}{\begin{tabular}{llllllc}  
\scriptsize
layer & $\tau$ &  $\eta$ & $\beta$ & $v_{th}$ & initial weights \\ 
\hline
Convolutional & 80  & 0.001 & 1 & 5 & [0, 2] \\
Hidden & $80$ &  0.01 & 1 & 50 & [0, 0.25] \\
Output & $80$ &  0.001 & 1 & 10 & [0, 0.5] 
\end{tabular}}
\end{center}
\end{table}

In Table~\ref{tab2}, we compare the STiDi-BP with some recent reported results which used supervised learning algorithms. 
As shown, \cite{R11,R13} used CSNN structure and achieved the highest performance, while, they are based on rate coding which has great deal of computation.
In the area of single-spike-timing-based supervised learning algorithms 
\cite{R18,S1}, and this work are the only implementation of CSNN and other implementations \cite{R8,R9,R10,R3} are fully connected networks.
In \cite{R3}, we introduced STiDi-BP algorithm and achieved the accuracy of $97.4\%$ with the network structure of $784-350-10$. 
While, other fully connected SNNs\cite{R8,R9,R10,R18} employed the traditional temporal BP which requires backward recursive gradient computation.
Zhang et al. in \cite{R18} employed rectified linear PSP based spiking neuron models and developed two SNNs. They reached the accuracy of $98.5\%$ for a fully connected network and $99.4\%$ for CSNN with the structure of $28*28-16C5-P2-32C5-P2-800-128-10$. 
Zhou et al.~\cite{S1} use IF neuron models with exponential decaying function and defined a direct relation between neuron’s pre and postsynaptic firing times same as \cite{R8}. They acheived $99.3\%$ accuracy for CSNN with the structure of $28*28-32C5-P2-16C5-P2-10$. 
Here we apply STiDi-BP to an R-CSNN architecture and acheive the state-of-the-art accuracy with the lower number of convolutional and hidden layers.

\begin{table*}  
\begin{center}
\caption{The classification accuracies of recent supervised SNNs with direct training on the MNIST dataset with some details such as input coding scheme and the network structure are provided in the table. The convolution layer and pooling layer are represented by C and P, respectively and layers are separated by -.} \label{tab2}
\begin{tabular}{llllc}  
\scriptsize
Model & structure & Coding &  Accuracy($\%$). \\
\hline
Mostafa (2017)~\cite{R8}& 784-800-10 & Temporal & 97.2 \\
Comsa et al. (2019)~\cite{R9}& 784-340-10 & Temporal & 97.9 \\
Kheradpisheh et al. (2020)~\cite{R10}& 784-400-10 & Temporal & 97.4 \\
Mirsadeghi et al. (2020)~\cite{R3}& 784-350-10 & Temporal & 97.4 \\
Zhang et al.(2020)~\cite{R18}& 784-800-10 & Temporal & 98.5 \\
W.Zhange et al. (2020)~\cite{R11}& 15C5-P2-40C5-P2-300-10 & rate & 99.5 \\
Fang et al.(2020)~\cite{R13}& 128C3-P2-128C3-P2-2048-100-10  & rate & 99.6 \\
Zhang et al.(2020)~\cite{R18}& 16C5-P2-32C5-P2-800-128-10 & Temporal & 99.4 \\
Zhou et al.(2020)~\cite{S1}& 32C5-P2-16C5-P2-10 & Temporal & 99.3 \\
STiDi-BP in R-CSNN (This paper)& 40C5-P2-1000-10 & Temporal & 99.2 
\end{tabular}
\end{center}
\end{table*}

\begin{figure}[ht]
\centering
\includegraphics[width=0.5\textwidth]{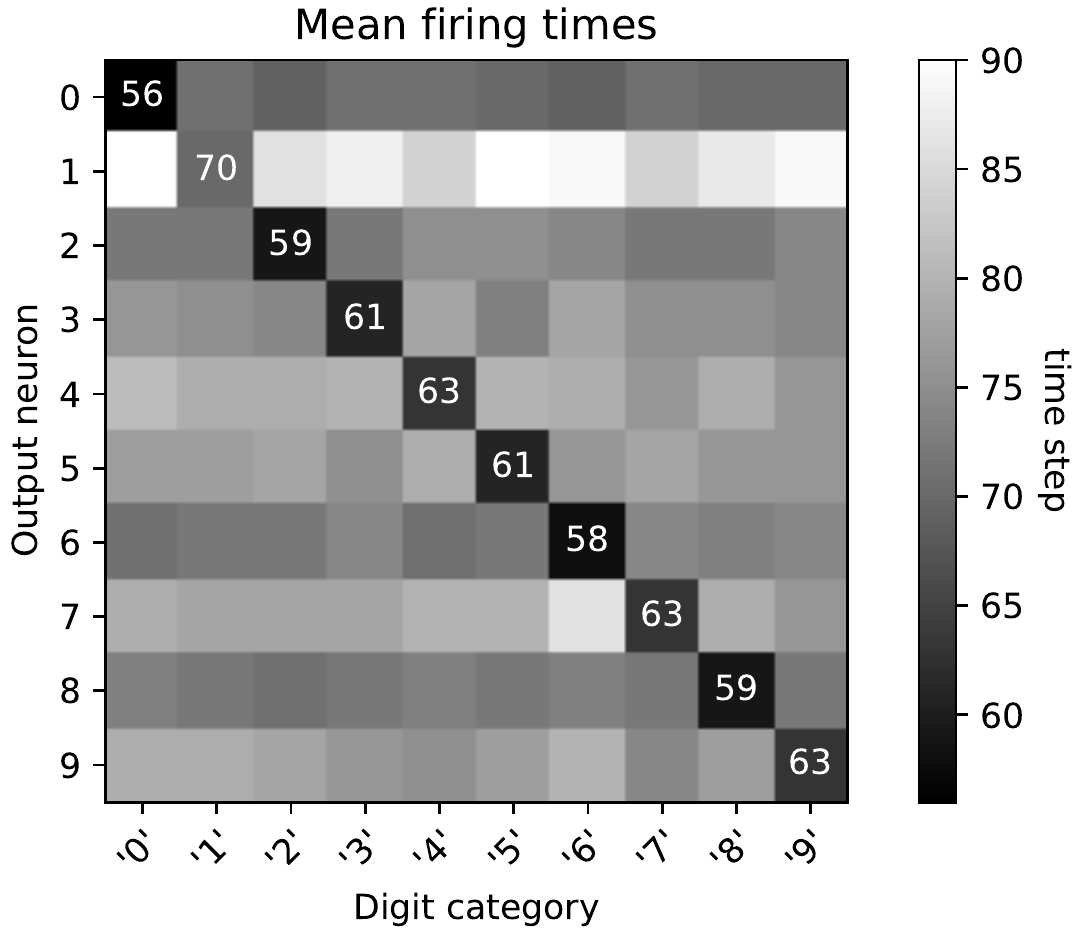}  
\caption{The mean firing time of each output neuron (rows)
over the images of different digit categories (columns) in R-CSNN.} \label{FIG3}
\end{figure}

\begin{figure}[ht]
\centering
\includegraphics[width=0.5\textwidth]{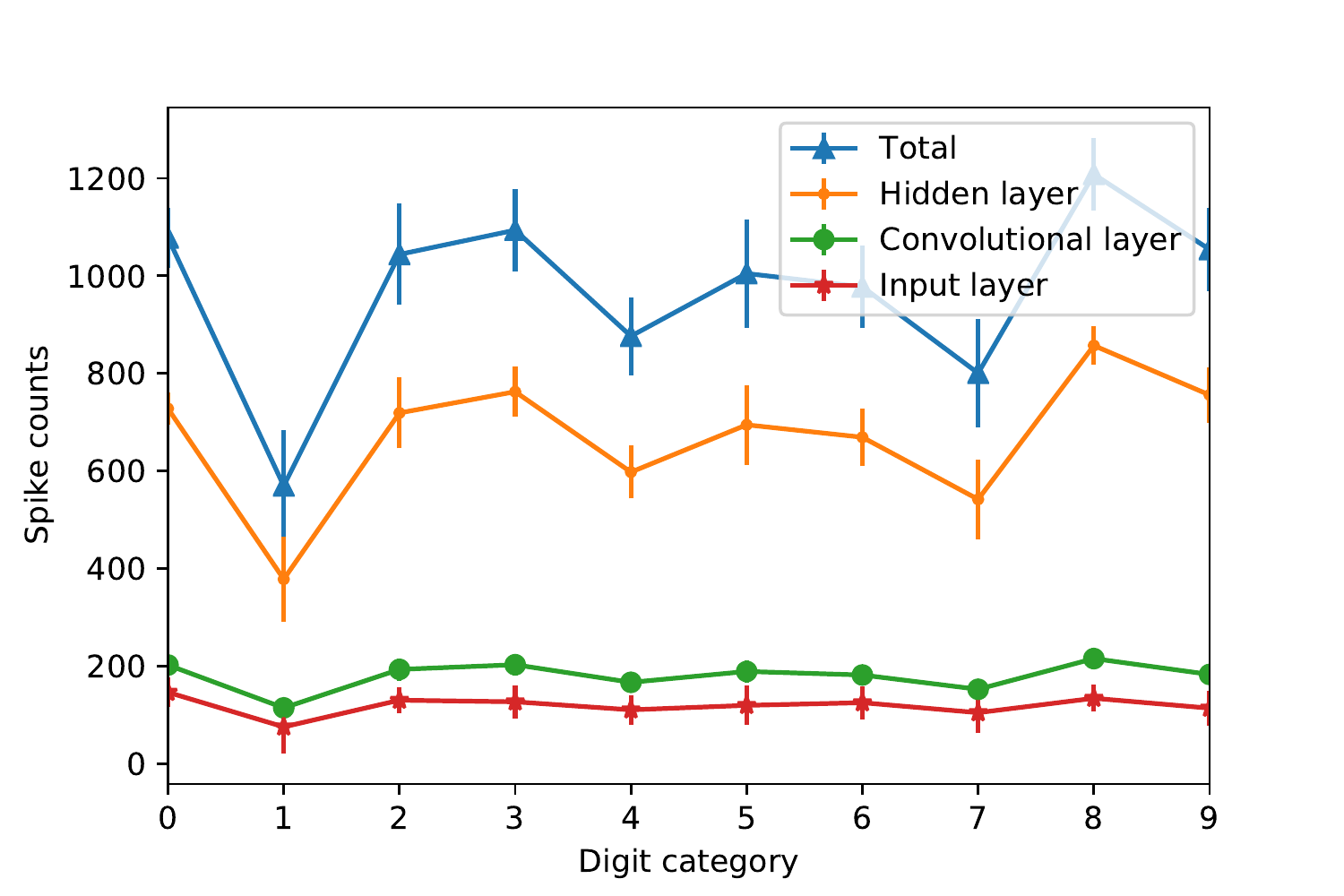}  
\caption{The mean required number of spikes in the input, convolutional, hidden, and total layers in R-CSNN.} \label{FIG4}
\end{figure}

The mean firing time of each output neuron over the images of different categories and the mean required spikes of all layers are depicted in Figure~\ref{FIG3} and Figure~\ref{FIG4}, respectively.

According to Figure~\ref{FIG3}, Each output neuron tends to fire earlier for images of its corresponding category which confirms that to recognize each input digit, it is not necessary to give all its input spikes to the network. 
Here digit $1$ has the maximum mean firing time because it covers the pixels that are common among most other digits. Therefore, the network needs much longer time to detect it.
On the other hand, Figure~\ref{FIG4} shows that the network is able to detect the class corresponding to the input image by firing a limited number of neurons in each layer which helps to make very rapid decisions about the input categories.
For example, the network needs only $590$ spikes in total to correctly recognize digit $1$, which has the maximum mean firing time of $70ms$.
And, the maximum number of mean required spikes is related to the digit $9$, which is only $1198$ spikes.

These two properties (that are illustrated in Figure~\ref{FIG3} and Figure~\ref{FIG4}), are the most important reasons for low cost and high computational speed in single-spike-based temporal SNNs.

\begin{figure}
\centering
\includegraphics[width=0.5\textwidth]{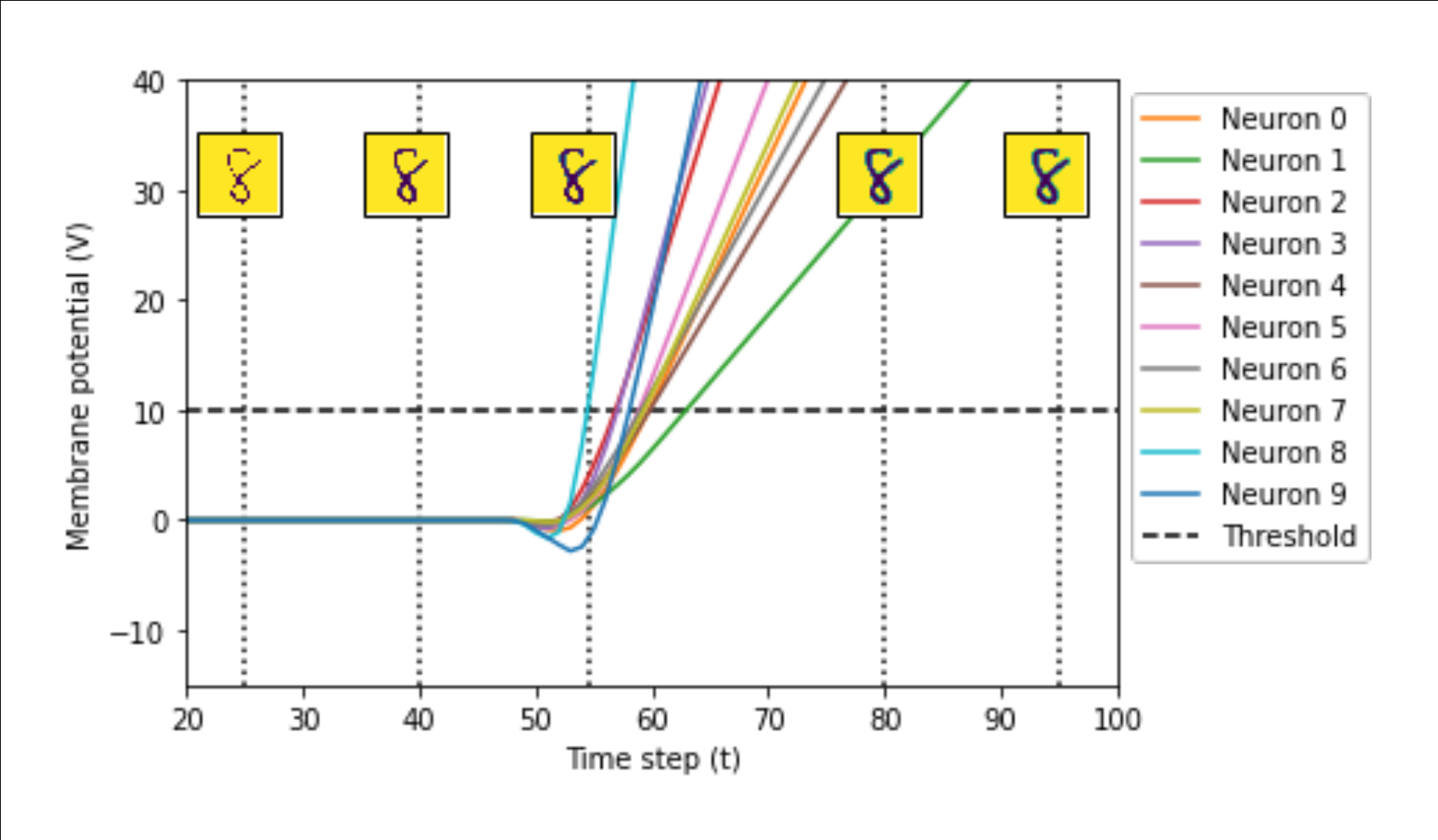}  
\caption{The trajectory of the membrane potential of all output neurons for sample $8$  test image. The incoming input spikes up to the time step $54$ contribute to the digit classification and the remaining spikes are ignored.} \label{FIG5}
\end{figure}

This  is shown more clearly in Figure~\ref{FIG5}, where, the membrane potential of output neuron for a sample $8$ test image and the accumulated input spikes until the $25$, $40$, $54$, $80$, and $95$ time steps are depicted.
As soon as the membrane potential of an output neuron reaches the threshold, the network assigns the corresponding class to the input image and can stop the computations. 
Here, the membrane potential of $8^{th}$ output neuron overtakes others and reaches the threshold at time step $54$. 
As seen, there is no need to propagate all input spikes to determine the category of the input image. By propagating a limited number of input spikes up to the $54^{th}$ time step, the membrane potential of the correct output neuron crosses the threshold and the network can classify the input image.

\subsubsection{Fashion-MNIST dataset}

Fashion-MNIST is a dataset of Zalando's article images~\cite{fashion} which has the same image size and structure of training and testing splits as MNIST, but it is more challenging classification problem.

Here we develop an R-CSNN with the structure of $20C5-P2-40C5-P2-1000-10$.
The maximum simulation time is $T_{max}=100$ and other parameters of each layer are listed separately in Table~\ref{tab3}.

\begin{table}
\begin{center}
\caption{Model parameters for Fashion-MNIST dataset in R-CSNN.} \label{tab3}
\resizebox{0.5\textwidth}{!}{\begin{tabular}{lllllc}
\scriptsize
 layer & $\tau$ &  $\eta$ & $\beta$ & $v_{th}$ & initial weights \\
\hline
convolutional$^1$& 80  & 0.0001 & 1 & 5 & [0, 2] \\
convolutional$^2$& 80  & 0.001 & 1 & 10 & [0, 1] \\
$Hidden$& $80$ &  0.1 & 1 & 100 & [0, 1] \\
$Output$& $80$ &  0.01 & 1 & 50 & [0, 1] \\
\end{tabular}}
\end{center}
\end{table}

The classification accuracies and characteristics of different approaches on Fashion-MNIST dataset are shown in Table~\ref{tab4}.

\begin{table*}
\begin{center}
\caption{The classification accuracies of recent supervised SNNs with direct training on the fashion-MNIST dataset with input coding scheme and the network structure are provided in the table. The convolution layer and pooling layer are represented by C and P, respectively and layers are separated by -.} \label{tab4}
\begin{tabular}{llllc}
\scriptsize
Model & structure & Coding &  Accuracy($\%$). \\
\hline
Kheradpisheh et al. (2020)~\cite{R10}& 784-1000-10 & Temporal & 88.0 \\
Zhang et al.(2020)~\cite{R18}& 784-1000-10 & Temporal & 88.1 \\
W.Zhange et al. (2020)~\cite{R11}& 32C5-P2-64C5-P2-1024-10 & rate & 92.8 \\
Fang et al.(2020)~\cite{R13}& 128C3-P2-128C3-P2-2048-100-10   & rate & 93.8 \\
Zhang et al.(2020)~\cite{R18}& 16C5-P2-32C5-P2-800-128-10 & Temporal & 90.1 \\
STiDi-BP in R-CSNN (This paper)& 20C5-P2-40C5-P2-1000-10 & Temporal & \textbf{92.8}
\end{tabular}
\end{center}
\end{table*}

\cite{R11, R13} that achieve the highest performance with the convolutional structure, are based on rate coding scheme with the great deal of computation.
Among the temporal coding-based SNN approaches, \cite{R18} and this work are the only implementation of CSNN in which, the proposed learning algorithm outperforms and reaches the accuracy of $92.8$ which is not much different from rate coding schemes.

\begin{figure}
\centering
\includegraphics[width=0.5\textwidth]{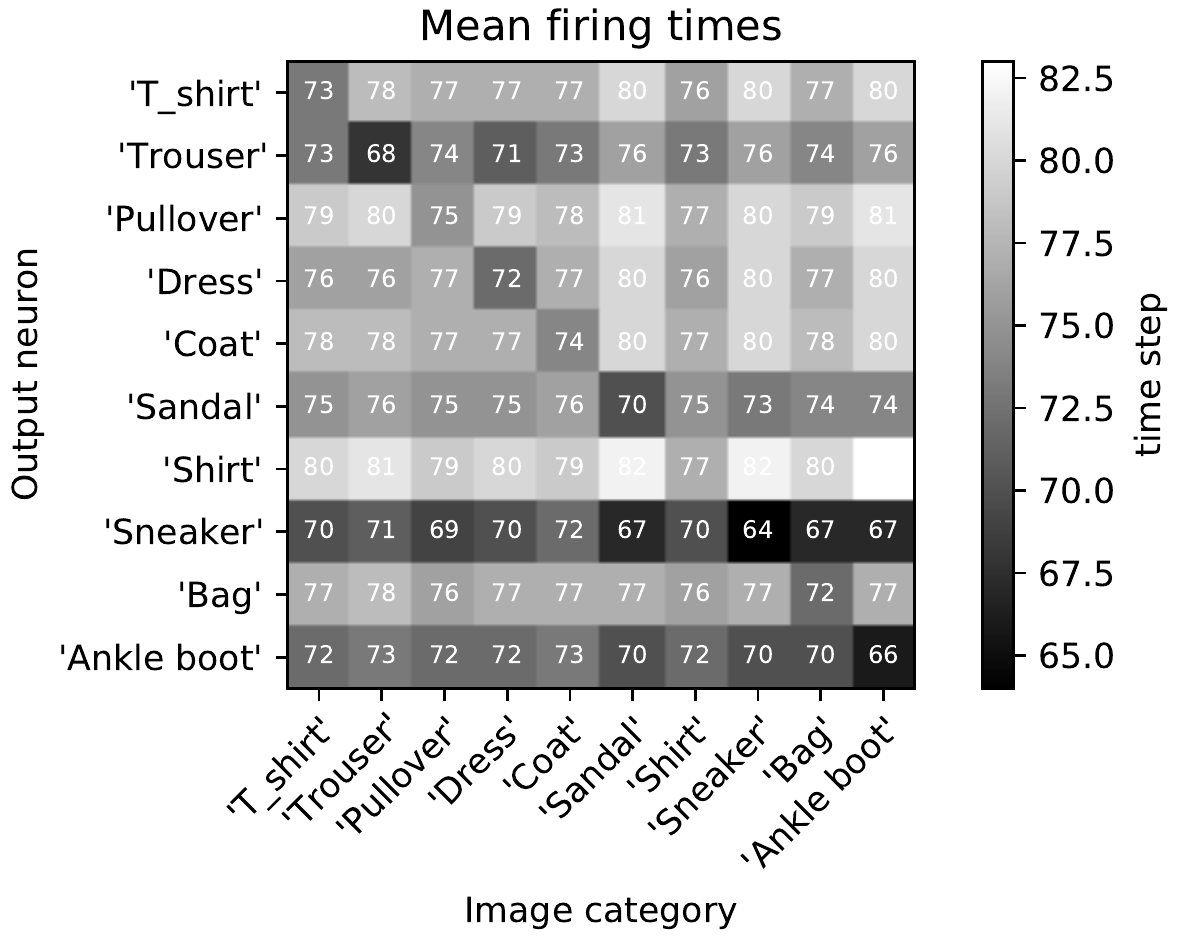} 
\caption{The mean firing times of the output neurons over the Fashion-MNIST categories in R-CSNN.} \label{FIG66}
\end{figure}

The mean firing times of the output neurons for each categories of Fashion-MNIST are illustrated in Figure~\ref{FIG66}.
The correct output neuron tends to fire earlier than the others for its corresponding category, yet the difference between the mean firing times of the correct and some other output neurons is small. This is due to the similarities between images of different categories in Fashion-MNIST compared to MNIST.

\begin{figure}[ht]
\centering
\includegraphics[width=0.5\textwidth]{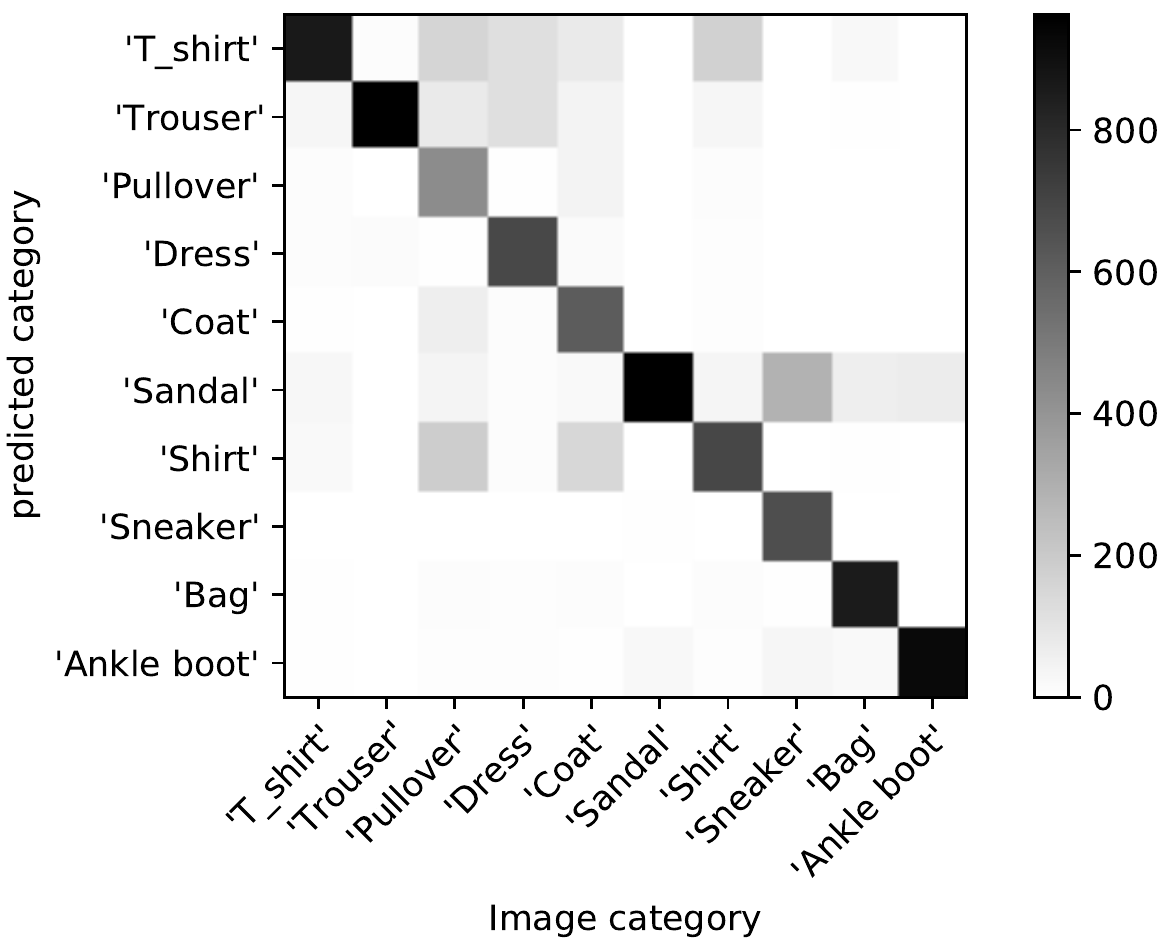} 
\caption{The confusion matrix of R-CSNN on Fashion-MNIST.} \label{FIG6}
\end{figure}

These similarities are more clearly shown in Figure~\ref{FIG6}, where, the confusion matrix of the proposed learning algorithm on Fashion-MNIST is depicted.
According to Figure~\ref{FIG6}, the network confuses T\_shirt, shirt, dress, coat and pullover due to their similar images. And, the same goes for ankle boots, sandals, and sneakers that have close mean firing times together. 
For example, the output neuron corresponding to the `Ankle boot' sample has the mean firing time of $66$, while, it fires respectively at $70$ for `sneaker' and `sandal' samples, which are very close to $66$.

In Table~\ref{tab5}, we show the mean firing time of correct output neurons and the mean required number of spikes in all layers. 
Again, it is not necessary to give all the input spikes of an image to the network and by firing a limited number of neurons in each layer, the network can recognize the image category.

\begin{table*}
\begin{center}
\caption{The mean firing time (MFT) of the correct output neuron and the mean required number (MRN) of spikes in all the layers for each category of Fashion-MNIST in R-CSNN.} \label{tab5}
\begin{tabular}{lllllllllllc}
\scriptsize
Category & T\_shirt & Trouser &  Pulloiver & Dress & Coat & Sandal & Shirt & Sneaker & Bag & Ankle boot \\
\hline
MFT & 73 & 68 & 75 & 72 & 74 & 70 & 77 & 64 & 72 & 66 \\
MRN & 2836 & 2096 & 3227 & 2386 & 3169 & 1886 & 3020 & 1771 & 3015 & 2592
\end{tabular}
\end{center}
\end{table*}

\subsection{Binary weights}  
\subsubsection{MNIST dataset}
Here we apply STiDi-BP to directly train a CSNN with binary synaptic weights (B-CSNN) ${-1,1}$ and evaluate it on the MNIST dataset. 
The network has the same structure and parameters as R-CSNN in section $6.1.1$, except that the scaling factor (SCF) of each layer as another trainable parameter has been added to it. 
The parameter settings are provided in Table~\ref{tab6}.
In the convolutional layer, the value of the parameter $\mu$ is shared between all the weights.

\begin{table}
\begin{center}
\small
\caption{Model parameters for MNIST dataset in B-CSNN.}  \label{tab6}
\resizebox{0.5\textwidth}{!}{\begin{tabular}{lllc }
 layer & $\mu$ & initial SCF \\
\hline
convolutional & 0.0001 & [0, 2] \\
Hidden &  0.001 & [0, 3] \\
Output & 0.0001 & [0, 2] 
\end{tabular}}
\end{center}
\end{table}

\begin{figure}
\centering
\includegraphics[width=0.5\textwidth]{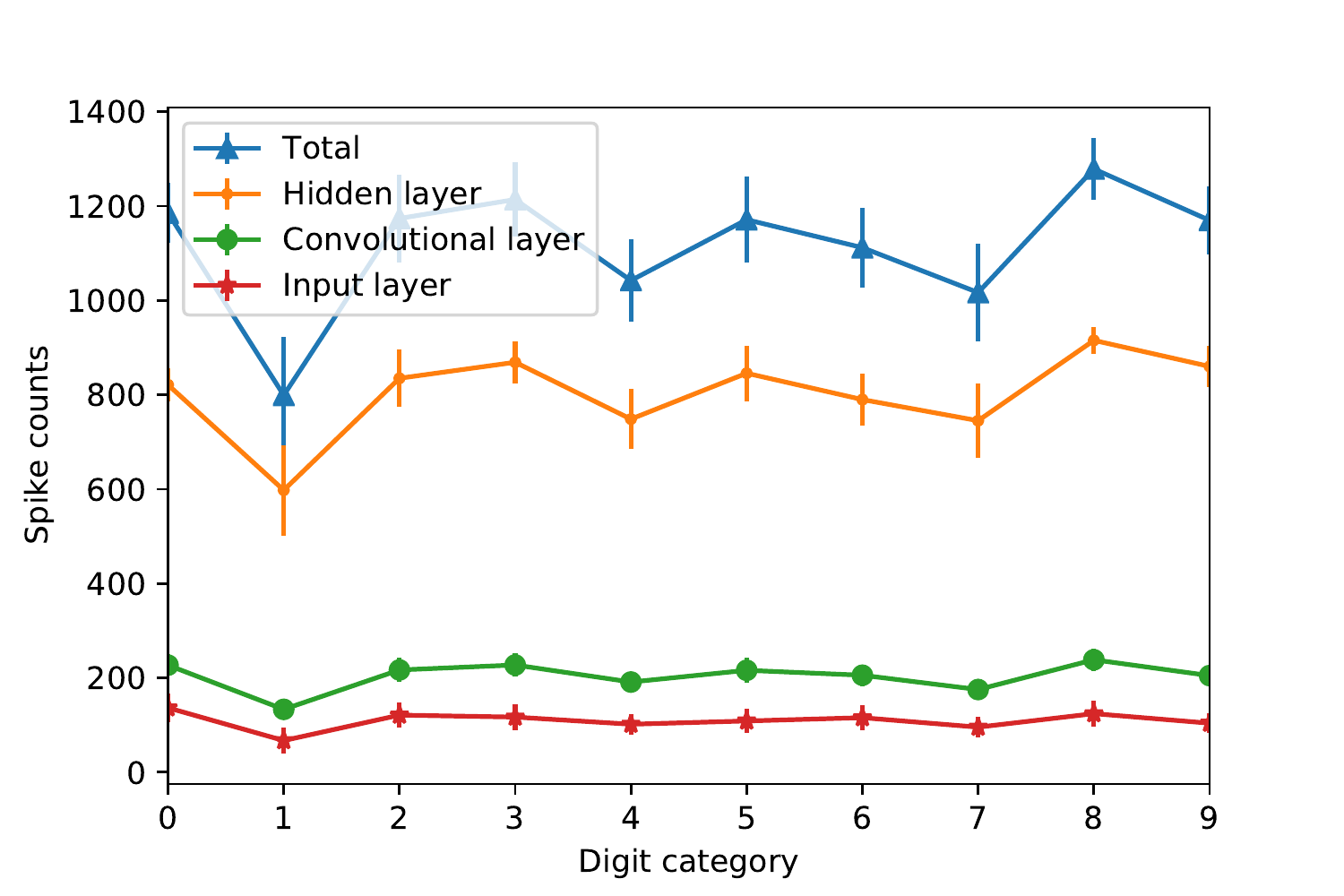}
\caption{The mean required number of spikes in the input, convolutional, hidden, and total layers in B-CSNN.} \label{FIG7}
\end{figure}

The mean required number of spikes in each layer is depicted in Figure~\ref{FIG7}. 

By comparing Figure~\ref{FIG7} with Figure~\ref{FIG4},
we see that the mean required number of spikes in the input and convolutional layers are almost equal to the R-CSNN and the difference is in the hidden layer, where, B-CSNN needs more number of spikes than R-CSNN.
In fact, due to the use of binary weights, the B-CSNN should wait for more time steps in the hidden layer to detect the corresponding category of an input image. Therefore more spikes are generated in the hidden layer. 
For example, the network needs about $810$ spikes in total layers to correctly recognize digit $1$, while, there was $590$ spikes in R-CSNN.

In Table~\ref{tab7}, we show the classification accuracy of STiDi-BP on two networks of R-CSNN and B-CSNN and compare with BS4NN\cite{R17}. As mentioned before, BS4NN is the only network aimed at directly training multi-layer temporal SNNs with binary weights.
Because of employing convolutional structure, B-CSNN outperforms the BS4NN. And, compared to R-CSNN, the performance of B-CSNN only dropped by $0.6\%$ which is due to the use of binary weights.

\begin{table}
\begin{center}

\caption{The classification accuracies of recent binary SNNs with direct training on the MNIST dataset.} \label{tab7}
\resizebox{0.5\textwidth}{!}{\begin{tabular}{lllc}
\scriptsize
 Model & structure &  Accuracy($\%$) \\
\hline
BS4NN\cite{R17} & 784-600-10 & 97.0 \\
R-CSNN &  40C5-P2-1000-10 & 99.2 \\
B-CSNN & 40C5-P2-1000-10 & 98.6 
\end{tabular}}
\end{center}
\end{table}

\subsubsection{Fashion-MNIST dataset}

Here we evaluate B-CSNN on the Fashion-MNIST dataset. The network has the structure of $20C5-P2-40C5-P2-1000-10$ with the initial scaling factors in range $[0, 10]$. The value of $\mu$ is $0.01$ for the convolutional layers and $0.1$ for the hidden and the output layers. In the convolutional layers, the value of $\mu$ is different for each convolutional filter and is trained independently of the others.
Other parameters are the same as Table~\ref{tab3}.

We illustrate the classification accuracy of the proposed learning algorithm on B-CSNN and R-CSNN in Table~\ref{tab8} and compare them with the BS4NN.

\begin{table}
\begin{center}
\caption{The classification accuracies of recent binary SNNs with direct training on the Fashion-MNIST dataset.} \label{tab8}
\resizebox{0.5\textwidth}{!}{\begin{tabular}{lllc }
\scriptsize
 Model & structure &  Accuracy($\%$) \\
\hline
BS4NN\cite{R17} & 784-1000-10 & 87.3 \\
R-CSNN & 20C5-P2-40C5-P2-1000-10 & 92.8 \\
B-CSNN & 20C5-P2-40C5-P2-1000-10 & 92.0 
\end{tabular}}
\end{center}
\end{table}

As seen, B-CSNN outperforms the BS4NN, due to the use of convolutional structure. And, there is only $0.8\%$ dropped compared to R-CSNN. 
Therefore, the proposed algorithm is able to directly train a SNN with binary weights without any significant drop in the performance of the network.

The mean firing time of correct output neurons along with the mean required number of spikes of all layers are depicted in Table~\ref{tab9}.
As seen, the mean required number (MRN) of spikes of all layers in the B-CSNN is more than R-CSNN for each category. This can be due to the use of binary weights which makes B-CSNN to wait for more time steps to detect the corresponding category of an input image.

\begin{table*}
\begin{center}
\caption{The mean firing time (MFT) of the correct output neuron and the mean required number (MRN) of spikes in all the layers for each category of Fashion-MNIST in B-CSNN.}  \label{tab9}
\begin{tabular}{lllllllllllc}
\scriptsize
Category & T\_shirt & Trouser &  Pullover & Dress & Coat & Sandal & Shirt & Sneaker & Bag & Ankle boot \\
\hline
MFT & 74 & 70 & 74 & 73 & 72 & 71 & 75 & 68 & 71 & 67 \\
MRN & 3013 & 2255 & 3418 & 2541 & 3351 & 2023 & 3204 & 1824 & 3201 & 2742 
\end{tabular}
\end{center}
\end{table*}

\section{Discussion}
In this paper, we used a convolutional SNN, as the deep structure of SNN, with two modes of real-valued and binary weights. Then, we employed the proposed supervised learning algorithm, STiDi-BP, to directly train both networks. 
In the learning phase, we applied gradient descent (GD) to each layer independently to discard the backward recursive gradient computation. 
Therefore, the desired firing times at the middle layers should be defined by using presynaptic spike time displacement, while, the desired output spike times were defined by using the relative timing of output neurons.

The most important advantage of our proposed approach is the use of temporal single-spike coding. In such methods, calculations in the backward direction are only performed at the actual firing times and it is not required to backpropagate the error in all the time steps which decreases the computational cost and the required storage space.
The space complexity in each layer during the backward pass is $O(N)$. While, in the rate coding schemes, the space complexity in each layer in the backward pass is $O(NT)$, where $T$ is the number of time steps, due to backpropagating the error in all the time steps.
Also, contrary to the rate-based CSNNs, the max-pooling operation can be simply done by propagating the first spike emerging inside the receptive window of each pooling neuron. 

Many of the existing supervised learning algorithms are based on rate or multi-spike coding~\cite{B6,B7,B9,B10,B11,B12,B13} which require expensive computation.
There are few works that focus on the single-spike-based temporal coding~\cite{R3,R7,R8,R9,R10,R17,R18,S1}.
Among the existing single-spike-based temporal approaches, Zhang et al.~\cite{R18} and Zhou et al.~\cite{S1} are the only implementation of a convolutional SNN architecture with single-spike-timing-based supervised learning algorithms and others are based on fully connected networks.

With the exception of \cite{R18,S1}, the other CSNNs have been presented in two forms:
1- the converted version of traditional CNNs~\cite{R4,R5,R6,R46} and,
2- CSNNs that use rate coding or multi-spike-based coding schemes to be directly trained by BP~\cite{R11,R12,R13}.
These approaches are computationally expensive due to the use of rate coding scheme and are based on backward recursive gradient computation.

Experimental results confirmed that the proposed approach can be applied to a CSNN and it achieves acceptable results compared to \cite{R18}.
The CSNN trained by this algorithm reaches $99.2\%$ accuracy on MNIST dataset and accuracy of $92.8\%$ on Fashion-MNIST as the more challenging dataset.

Adapting the proposed learning rule to CSNN, removes backward recursive gradient computation, and reduces the complexity of neural processing and computational cost.

Binarizing the synaptic weights is another important improvement which helps optimization in hardware implementations of deep SNNs~\cite{A13, A43}.
Current Binary SNNs are the converted version of pre-trained BANNs~\cite{R14,R15,R16,R19}.
They train a BANN by using traditional BP and then, convert it into the equivalent BSNN with rate-based neural coding. 
Here, we developed a CSNN with binary weights, which are the sign of real-valued weights, and employed the proposed learning rule to directly train it. 
The forward pass is done with the binary weights and, in the backward pass, we updated the real-valued weights.
The proposed BCSNN uses single-bit of memories for implementing binary synapses and employs only one full-precision scaling factor in each layer or each convolutional filter. Therefore, the network size can be reduced by $32\times$ compared to a network with $32$-bit floating-point synaptic weights~\cite{R18}.
Also, due to the use of single-bit synapses, the multiplier blocks that impose high load of floating-point computation to the network can be replaced by one unit increment and decrement blocks~\cite{R18}.
The evaluation results shows that the BCSNN has a negligible performance drop compared to the CSNN, respectively $0.6\%$ and $0.8\%$ accuracy drop on MNIST and Fashion-MNIST datasets. While it has more advantages than the CSNN in terms of hardware implementation.
To the best of our knowledge, this is the first implementation that aim to directly train a deep structure of single-spike-based temporal SNNs with binary synaptic weights.
However, one of the most important challenges we face is to make the network deeper to solve more complex problems such as CIFAR10 or ImageNet classification which can be our future topic.

\begin{singlespace}

\end{singlespace}
\end{document}